\documentclass[letterpaper, 10 pt, conference]{ieeeconf}

\IEEEoverridecommandlockouts                              
\usepackage{times}
\usepackage{helvet}
\usepackage{courier}

\usepackage{latexsym}
\usepackage[english]{babel}
\usepackage{graphicx}
\usepackage{verbatim}

\usepackage{amsmath}
\usepackage{amsfonts}
\usepackage{relsize}

\usepackage{amsmath} 
\usepackage{amssymb}  
\usepackage{amsfonts}

\usepackage{algorithm}
\usepackage{algpseudocode}
\usepackage{epstopdf}

\usepackage{longtable}
\usepackage{textcomp}
\usepackage{url}
\usepackage{alltt}
\usepackage{caption}
\usepackage{subcaption}
\usepackage{graphicx}

\usepackage[table,dvipsnames,svgnames]{xcolor}

\newcommand{\needlegame}{\textit{Needle Master}}

\newcommand{\myeq}[2]{\begin{equation}\label{eq:#1}\begin{aligned}#2\end{aligned}\end{equation}}
\newcommand{\myeqref}[1]{\eqref{eq:#1}}

\newsavebox{\allttbox}

\begin{document}
%
\title{Towards Robot Task Planning From Probabilistic Models of Human Skills}
\author{Chris Paxton, Marin Kobilarov, and Gregory D. Hager\\
Johns Hopkins University\\
3400 North Charles Street\\
Baltimore, MD\\
}
\maketitle
\begin{abstract}
\begin{quote}
  We describe an algorithm for motion planning based on expert demonstrations of a skill.
  In order to teach robots to perform complex object manipulation tasks that can generalize robustly to new environments, we must (1) learn a representation of the effects of a task and (2) find an optimal trajectory that will reproduce these effects in a new environment.
  We represent robot skills in terms of a probability distribution over features learned from multiple expert demonstrations.
  When utilizing a skill in a new environment, we compute feature expectations over trajectory samples in order to stochastically optimize the likelihood of a trajectory in the new environment.
  The purpose of this method is to enable execution of complex tasks based on a library of probabilistic skill models.
  Motions can be combined to accomplish complex tasks in hybrid domains.
  Our approach is validated in a variety of case studies, including an Android game, simulated assembly task, and real robot experiment with a UR5.
\end{quote}
\end{abstract}

\section{Introduction}
\begin{figure}[bt!]
  \centering
  \includegraphics[width=\columnwidth]{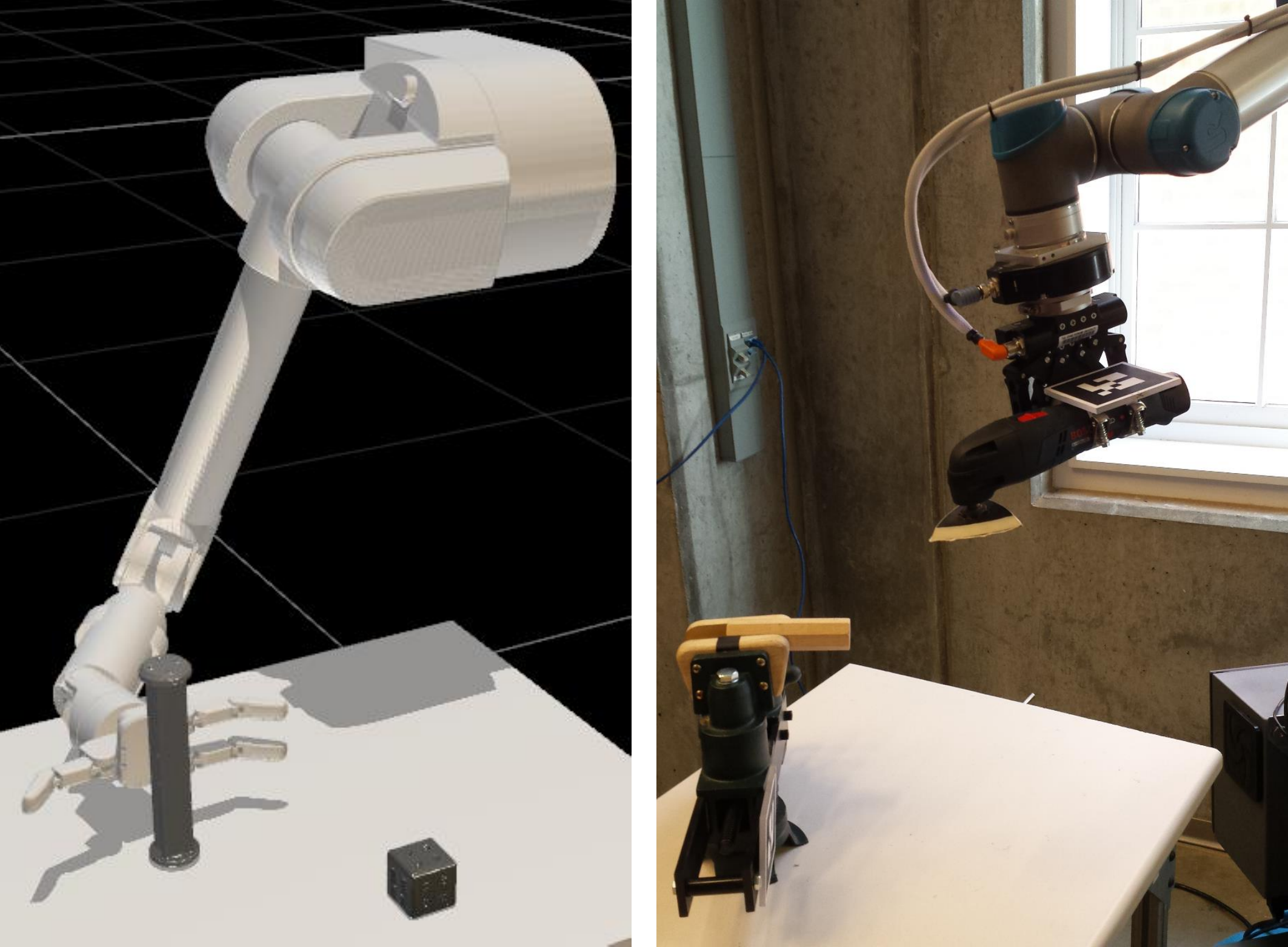}
  \caption{Simulated WAM arm performing an \texttt{APPROACH} action using the proposed method (left), and UR5 lifting a sander as a part of a learned task (right).}
  \label{fig:sim-approach}
\end{figure}

Many interesting robotic tasks involve multiple steps and substantial environmental variability.
Further, in a constrained environment, many steps are shared across tasks.
Consider the example task of assembling a structure out of different magnetic pieces, a simple version of which is shown in Fig.~\ref{fig:sim-approach}(left). The robot must be equipped with skills to
successfully approach each of the pieces,
close the gripper around each of the different pieces,
latch pieces together,
release and disengage from the pieces without causing damage to the structure.

We argue that successfully reproducing a task relies on (a) breaking tasks into re-usable, semantically meaningful skills, (b) learning representations of each of these skills, and (c) assembling and executing these skills in new environments.
In this paper we focus on assembling and executing robot skills
based on a probabilistic representation of those skills learned from demonstrations by an expert user.
Our primary goal is the development of a strategy for motion planning individual skills based on expert demonstrations, in order to enable development of a task-level planning algorithm using these skills.
These skills must represent tasks with complex dynamics and discrete state changes, such as grasping and manipulating objects.

The problem of learning a model of a task based on such a user demonstration is referred to as imitation learning. 
While reinforcement learning methods have been developed that allow us to adapt goal-directed robot skills to new environments~\cite{theodorou2010reinforcement,pastor2011skill,kober2011reinforcement,stulp2012path}, we focus on more general skill learning to adapt to new environments with a more abstract representation of goals.
Our approach uses a set of user demonstrations to learn a set of soft constraints on actions, and finds a path between different high-level goals while adhering to these constraints and minimizing collisions.

\begin{figure}[bt!]
  \centering
  \includegraphics[width=0.8\columnwidth]{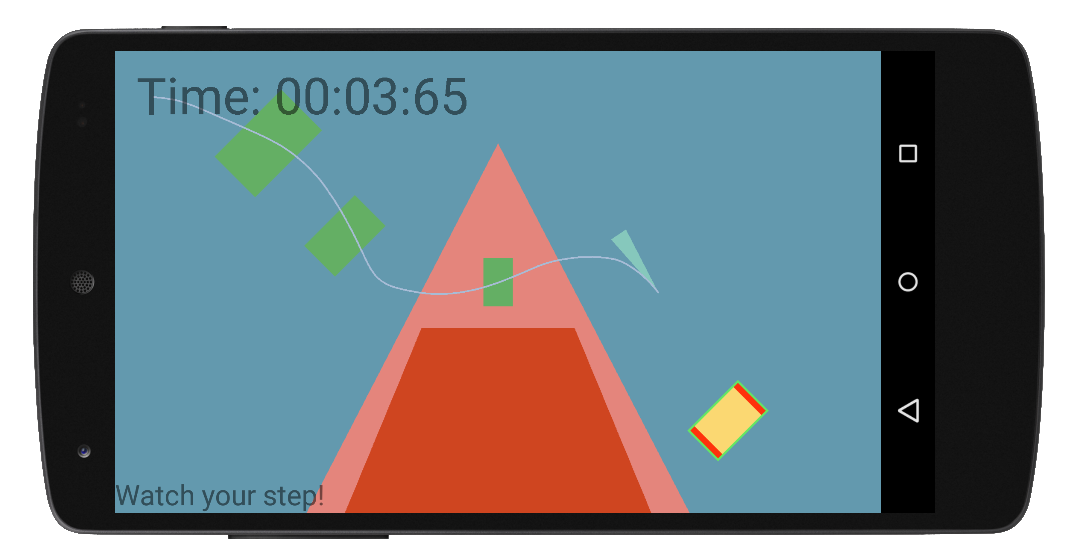}
  \caption{An example of a complex Needle Master level, showing different types of goals and obstacles imposing different constraints on the motion of the needle.}
  \label{fig:needlemaster}
\end{figure}

A standard approach to formulate a task is as a sequence of low-level actions, each associated with predicates, preconditions, and effects. A task $T$ is defined as a sequence of actions $A_i$ according to $T = \left\{A_i\right\}_{i=0}^{N_{\small T}}$ where each action is defined 
in terms of its specific preconditions and effects.
To execute action $A_i$, the effects of $A_1,\dots,A_{i-1}$ must have resulted in the preconditions of that action.
Symbolic planners allow us to reason about how to complete complex tasks. These planners have been combined with motion planning to allow robots to solve tasks that were previously impossible, as in~\cite{plaku2010sampling}.
However, engineering a system that allows a robot to use this sort of high-level knowledge is very challenging, and inevitably will require placing specialized domain knowledge that must be constantly updated. 

Instead, it would be advantageous to ground the definition of each possible state based on user demonstrations.
Any such grounding should allow robust motion planning in novel environments to accomplish low-level actions.
Some previous work has attempted to solve this problem through segmentation into lower-level action models~\cite{grollman2010incremental,butterfield2010learning,niekum2012learning} or through learning predicate conditions and effects~\cite{ahmadzadeh2015learning}.
However, these approaches do not consider the difficulties inherent in reproducing tasks in complex novel environments.

We propose an approach aimed at overcoming these issues by combining a probabilistic representation of a skill learned from expert demonstrations with motion planning. Reproducing such skills in a new environment can be regarded as producing the same intended effects (represented by feature observations). However, it is difficult to determine what these intended effects are from observation.
We use a probabilistic model to capture the relationships between the robot and the environment that is expected to occur during an expert performance of a task,
wherein each skill is represented as a probability distribution over expected features that capture the intended effects of that skill.
These may be changes in the relationships between objects in the environment or between the robot and its environment, for example.

The contributions of this paper can be summarized as: (1) derivation of an algorithm for motion planning using probabilistic models of expert skills,
(2) preliminary approach for combining these learned skills for executing complex tasks,
and (3) experimental validation of this algorithm in a number of case studies, including the robotic tasks shown in Fig.~\ref{fig:sim-approach} and an Android game shown in Fig.~\ref{fig:needlemaster}.

\section{Background}

We seek to enable combined task- and motion-planning for complex, multi-stage tasks based on learned representations of those tasks.
Prior work exists in describing the relationship between high- and low-level actions and in learning representations of actions from demonstration, but does not combine learning with low-level motion planning.

Kr\"uger et al. provide a formal description of object-action complexes~\cite{kruger2009formal},
which describe behaviors in relation both to objects and their intended effects.
However, there still exists much work to do when it comes to coming up with task plans involving a sequence of movements.
These object-action complexes can be associated with learned low-level actions, and segmented based on predicates~\cite{wachter2013action}.
We likewise use predicates to provide the segmentation used to learn our models of low-level skills, but we do not use them during execution.

Gaussian Mixture Regression was first used in~\cite{calinon2007learning} to recover expected end effector and joint positions plus object distances over time, and then compute a path based on this information.

Dynamic Movement Primitives (DMPs) are another imitation learning method that has proven useful for modeling low-level actions. 
DMPs model motions as a set of dynamical systems~\cite{schaal2006dynamic}.
Pastor et al. used reinforcement learning with DMPs and multiple human demonstrations to learn a model of expected features when executing two robotic tasks in~\cite{pastor2011skill}: shooting pool and flipping over a box with a pair of chopsticks.
Kormushev et al. used a modified version of DMPs together with reinforcement learning to adapt to new environments with a known goal~\cite{kormushev2010robot}.


Otherwise, DMPs are often adapted to new environments through reinforcement learning approaches such as Path Integral Policy Improvement, first proposed in~\cite{theodorou2010reinforcement}. This method has been applied to motion planning with DMPs~\cite{pastor2011skill}.
Work by Kober et al. uses reinforcement to adapt to new situations by adapting latent variables~\cite{kober2011reinforcement}.
These methods for reinforcement learning were further expanded upon by Stulp et al., who proposed Path Integral Policy Improvement with Covariance Matrix Adaptation~\cite{stulp2012path}.
These techniques are also closely related to the Cross-Entropy Method for motion planning~\cite{kobilarov2011cem}.
One of the major differences between the aforementioned reinforcement learning techniques and the algorithm proposed in this paper is that we do not know the correct goal state for our system, and our method is specifically derived for a probabilistc setting.
Our algorithm finds both an approximately optimal goal state and a plan for arriving at this state, expanding on the motion planning approach in~\cite{kobilarov2011cem}.

In~\cite{grollman2010incremental,butterfield2010learning,niekum2012learning} low-level action models were learned concurrently with a task model.
Grollman et al.~\cite{grollman2010incremental} learn a finite state machine task model as an infinite mixture of Gaussian experts,which is then applied to a robot soccer task.
In~\cite{butterfield2010learning} the authors performed a similar task, using an HDP-HMM to prevent perceptual aliasing by modeling time dependencies between actions.
Niekum et al.~\cite{niekum2012learning} used Beta Process Autoregressor Hidden Markov Models (BP-AR-HMMs) to learn tasks given unstructured expert demonstrations for a number of object manipulation tasks. After segmenting using the BP-AR-HMM, they learned DMPs representing different action primitives.
In this paper, we do not use such an approach to automatically segment our data, instead using predicates as in~\cite{wachter2013action}.
However, these papers do not perform any motion planning, which limits their ability to adapt to new environments. 

In~\cite{ahmadzadeh2015learning} the authors learn \emph{pull} and \emph{push} actions and associate them with changes in the state of the world in the form of PDDL predicates.
W\"achter et al. reproduce a sequence of actions from a human demonstration by mapping observations onto a library of Object-Action Complexes associated with preconditions and effects~\cite{wachter2013action}.

\section{Algorithm}

Each action $A$ in a given task is encoded as a probability distribution over a set of features that will be produced during a successful instantiation of a skill in a new environment. The features are denoted by $x\in\mathbb{R}^n$ and defined using the function $\phi$ through relationship
\[ 
x = \phi(t,s,u),
\]
where $t\in[t_0,t_f]$ denotes time, $s\in\mathbb{R}^d$ is the robot state, and $u\in\mathbb{R}^m$ are the applied control inputs. The model associated to each action $A$ is denoted by $p_D(x|A)$ and is computed using unsupervised learning from expert demonstrations, typically assuming a parametric density $p_D$. A joint model of a task $T$ consisting of multiple actions can be constructed using a density $p_D(x|T) \propto p_D(x|A_0)\cdots p_D(x|A_{n_T})$ assuming conditional independence between actions.


We then formulate the generalization of such a learned skill to a new environment as an optimization problem. This is accomplished by parameterizing trajectories using a parameter $\xi\in\mathcal Z$, where $\mathcal Z$ represents the space of all possible parameters resulting into valid trajectories. Since robot perception and motion are uncertain, each parameter induces a density $p(\tau|\xi)$ where
\[
  \tau=\{t_0,s_0,u_0,t_1,s_1,u_1,\dots,t_N,s_N,u_N\}
\]
denotes the system trajectory. For instance, $\xi$ would typically define a reference trajectory and an associated tracking control law resulting in the density
\[
p(\tau|\xi) = p(s_0)\prod_{i=0}^{N-1} p(s_{i+1} | s_i, u_i) p(u_i | s_i, \xi).
\]
In practice, given $\xi$ the trajectory $\tau$ will either be sampled using a high-fidelity simulator or generated by the real robot.




The aim of employing an optimization-based approach is to produce a feasible trajectory which optimally approximates the effects of an expert's action,
represented as a set of features capturing object-actor and object-object relationships in the workspace.
Therefore, the optimization cost is encoded as the likelihood of generating new features $x$ given the expert-derived probability distribution for the action,
i.e. with respect to the constructed model $p_D(x|A)$.

We propose to employ stochastic trajectory optimization to solve the motion planning problems in constrained environments~\cite{rubinstein2004cross,kobilarov2011cem}.
This is accomplished by introducing an artificial \emph{surrogate} distribution over $\mathcal V$ that will induce a distribution
over trajectories $\tau$ and over the corresponding features $x$ along these trajectories.
The surrogate will then be iteratively optimized until it becomes optimally close (in a distribution sense) to the expert density $p_D(x)$
without violating the constraints of the environment such as obstacles and joint limits.
The surrogate model is built using a parametric density $\pi(\xi|v)$ such as a multivariate Gaussian or a GMM with parameters $v$. 

More specifically, the cost is defined as the Kullback-Leibler divergence between the probability distribution of expected features and the average probability of observing those new features from dynamically feasible robot states.
It is given by:
\[
  J(\tau) = \dfrac{1}{N} \sum_{i=0}^N D_{\text{KL}}(p_D(x_i) || p(x_i | v) ),
\]
This is equivalent to minimizing the KL divergence for all observed states along a trajectory $\tau$.
For the $i$th observation in a given $\tau$ we use the notation:
\myeq{kl}{
D_{\text{KL}}(p_D(x_i) || p(x_i|v)) \triangleq \int p_D(x_i) \frac{\log p_D(x_i)}{\log p(x_i | v)}dx_i,
}
where $p(x_i|v) = \int I_{\{x_i = \phi(t_i,s_i,u_i)\}} p(\tau|\xi) \pi(\xi | v) d\tau d\xi$ is the probability density of a feature at time $t_i$ from the surrogate model $v$.

The optimal parameter $v^*$ can be obtained by noting that
\begin{align}
  & \ \ \min_v \dfrac{1}{N} \sum_{i=0}^N D_{\text{KL}}(p_D(x_i) || p(x_i | v) ), \\
  & =   \min_v \dfrac{1}{N} \sum_{i=0}^N \int - p_D(x_i) \log p(x_i | v) dx_i,
\end{align}
where $x_{i,j}=\phi(t_{i},s_{i,j},u_{i,j})$ is a generated feature from robot state $s_{i,j}$ at time $t_i$ along the sampled trajectory $\tau_j\sim p(\cdot | \xi_j)$ for $\xi_j\sim \pi(\cdot | v_0)$.


We can approximate this solution by drawing $M$ i.i.d. samples $\xi_1,\ldots,\xi_M$ from $v_0$:
\begin{align}
  & \approx \min_v \dfrac{1}{N} \sum_{i=0}^N \sum_{j=1}^M - p_D(x_{i,j}) \log\pi(\xi_j | v_0), \label{eq:minkl}
\end{align}

The necessary conditions for a minimum correspond to setting the gradient of~\eqref{eq:minkl} to zero, i.e. by solving the equality
\begin{align}
  \sum_{i=0}^N \sum_{j=1}^M - w_{i,j} \nabla_v \log \pi(\xi_j | v) = 0 \label{eq:optcond},
\end{align}
where the weights $w_{i,j}$ are given by $  w_{i,j} \triangleq p_D(x_{i,j}) p(x_j | \xi_i)$.
In practice, we often assume that $p(x_j | \xi_i) = 1$.
This is the case wherever system dynamics are deterministic.

When $\pi(\cdot |v) = \mathcal N(\cdot | \mu, \Sigma)\vert_{\mathcal V}$ (i.e. a single multivariate Gaussian with domain restricted to feasible parameter set $\mathcal Z$), the relationship~\eqref{eq:optcond} can be solved in closed form as
\begin{align}
\mu = \sum_{j=1}^M \bar w_{j} \xi_j, \quad \Sigma = \sum_{j=1}^M \bar w_{j} (\mu-\xi_j)(\mu-\xi_j)^T,
\end{align}
where $w_{j} = \sum_{i=0}^N w_{i,j}$ and $\bar w_j = w_j/\sum_{j=1}^M w_j$.
When $\pi(\cdot |v)$ is a GMM the minimization~\eqref{eq:minkl} is performed using a weighted expectation minimization (EM) algorithm. 

In practice, the optimal parameter $v$ is computed iteratively starting with some nominal choice $v_0$ which approximately covers the trajectory space of interest.
At each iteration we draw $M$ samples $\xi_j \sim \pi(\cdot|v_0), j \in 1,\dots,M$  and compute the next $v$ by minimizing~\eqref{eq:minkl}.
At the next iteration $v_0$ is set to $v$ and the process continues until the cost converges. 

\subsubsection{Executing a Task}

When attempting to complete a complex task, however, the approach outlined above is not sufficient. Optimizing individual actions may leave the robot in a situation where it cannot perform a necessary next step correctly.

One approach is to employ a joint model $p_D(x|T)$ over the whole task in place of $p_D(x|A)$. In such case it is expected that the parameter $\xi$ must encode a more complex trajectory which for instance could have discrete or binary variables corresponding to switching states (such as ``close-gripper'' or ``open-gripper'').
However, as the dimension and complexity of $\mathcal{V}$ grows, solving the optimal control problem becomes intractable.


Instead, we employ a simpler approach which still optimizes a trajectory $\tau$ over a single action $A_K$ with the exception that it also ensures smooth transition to the next action $A_{k+1}$. This is achieved by maximizing the likelihood of the final feature $x_N$ under action $A_{k+1}$.
To this end we employ $p_D(x_i|A_k)p_D(x_N|{A_{k+1})}$ in place of $p_D(x_i|A)$ in~\eqref{eq:minkl} leading to the new weights
\myeq{weights-task}{
        w_{ij} \triangleq p_D(x_{i,j}|A_k)p_D(x_{N,j}|A_{k+1}),
}
where $p_D(x|A_i)$ is the expert density of action $A_i$.



\subsubsection{Avoiding Obstacles and Joint Limits}

The cross-entropy method provides a straightforward way of dealing with obstacles and joint limits that may prevent us from reproducing a task in a new environment.
Rather than using potential fields for object avoidance as per~\cite{park2008movement}, we constrain $\mathcal{Z}$ to consist only of the space of valid trajectories.
This means that when drawing our $M$ samples, we remove samples currently in collision or past joint limits in our new environment and continue to draw sample trajectories until we have all $M$ valid examples.
This allows us to accomplish both motion planning and skill imitation through the same framework. This works effectively in practice as long as the task does not require generalization in environments with very narrow passages that the system has never been trained on. Such cases are extremely difficult since the probability of obtaining samples in the narrow passage is close to zero, unless an informative nominal density parameter $v_0$ is used with enough probability mass over such regions.

\subsubsection{Controlling Step Size}
In order to ensure we arrive at an optimal solution, we want to control the size of the steps we take during trajectory optimization.
To this end we introduce an extra parameter $0 < \alpha < 1$, which controls the size of steps taken at each iteration.
Given $\Sigma_i^*$ as the optimal $\Sigma$ at iteration $i$,
we compute $\mu_{i+1}$ and $\Sigma_{i+1}$ as:

\myeq{step-size}{
	\mu_{i+1} &= (1 - \alpha)\mu_i - \alpha \mu^*_i \\
	\Sigma_{i+1} &= (1 - \alpha)\Sigma_i - \alpha \Sigma^*_i
}

Since $\Sigma$ determines our search region, introducing this step size term lets us prevent our search from converging prematurely.
In practice we found setting this value to $0.50-0.75$ gave us the best results. 

\subsubsection{States with Low Covariance}
Our method should allow us to capture states that occupy a very narrow region of the possible feature space, meaning that a multivariate Gaussian or mixture of multivariate Gaussians representing this skill will have an extremely small covariance matrix.
To reduce this issue, we add a term to the diagonal entries in $\Sigma$ at each iteration. This fixed quantity is reduced at every iteration.
This simple strategy greatly increases the range of possible skills we can learn.

\section{Experiments}

We performed three experiments in different experimental domains representing a wide variety of tasks and potential applications of our method.
These tasks will be segmented into different actions via user-defined predicates, and the objects used in the predicates associated with each action will be used as parameters.
In the future, segmentation and feature selection might be performed automatically.


We developed an Android game to explore how user actions relate to different features of the environment and to allow us to test different approaches for modeling action primitives.
This game, called ``\needlegame{}'', is inspired by the suturing task common in robotic minimally invasive surgery.
It is available for Android smart phones and tablets, and can be downloaded for free from the Google Play Store\footnote{\url{https://play.google.com/store/apps/details?id=edu.jhu.lcsr.needlemaster}}.

In the \needlegame{} case study described in Section~\ref{sec:needle-master}, our objects are all \emph{gates}.
These are 2D poses defined by $(x,y,\theta)$ and having a height and a width.
Similarly, the simulated and real robot experiments use object poses to compute features.
In effect, these features are capturing soft constraints on the positions the needle or end effector can take based on the positions of objects that are important to the task.

In Section~\ref{sec:robots}, we examine several robotic tasks, both in simulation and in the real world. The different tasks we describe in this section show that our method can encode a broad range of robotic skills.

\subsection{Case Study: Needle Master}\label{sec:needle-master}

We initially explored our method in simulation environments with ``perfect'' sensing capabilities.

In \needlegame{}, users need to steer a 2D needle through a number of target gates positioned in patient tissue, as shown in Fig.~\ref{fig:needlemaster}.
These gates represent needle insertion points in a suturing task.
We aim to learn the way users respond to variations in the features of the environment, particularly in the position and orientation of these gates.
Maintaining the correct orientation of the needle relative to gates and tissue is very important, as is making a path that does not go into high-risk ``tissue'' regions.

Users can rotate the needle and control its velocity. The rules of the game are:
\begin{itemize}
  \item The needle must exit off of the right side of the screen.
  \item The needle must pass through gates in order. Hitting the top or bottom of the gate will ``break'' that gate, causing the user to score less points for passing through it.
  \item The needle must not hit dark red ``deep tissue'', representing vital organs.
  \item The needle must not cause too much damage to pink ``tissue''. Players can damage this tissue if they rotate the needle too much.
\end{itemize}

\begin{figure*}[bt!]
\centering
\includegraphics[width=1.8\columnwidth]{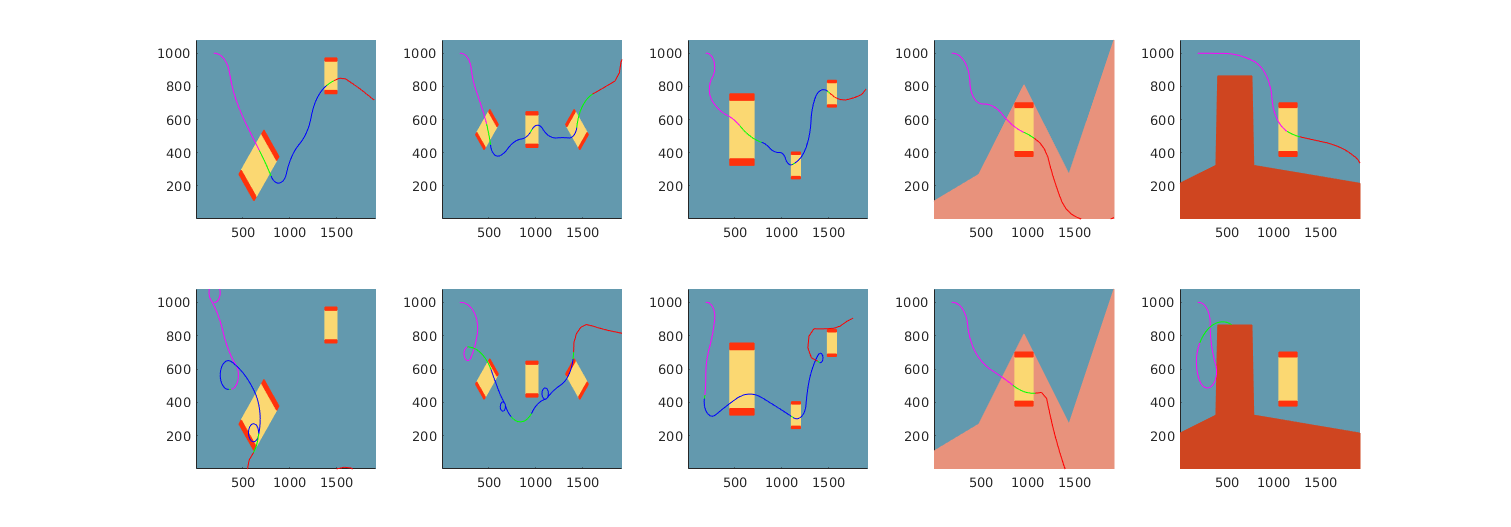}
\caption{Comparison showing our method on Needle Master levels $4-8$ (top) vs. a version of the model without goal term introduced in Eq.~\myeqref{weights-task} (bottom). Colors indicate which learned skill was active at each point in the trajectory: approaching a gate is purple, passing through a gate is green, connecting two gates is blue, and moving to the exit is red.}
\label{fig:nm-levels}
\end{figure*}

In effect, the users' goal is to come up with a motion plan that pass through as many gates as possible without breaking them, while avoiding red obstacles and constraining needle rotation in pink obstacles.
Fig.~\ref{fig:needlemaster} shows a complex environment users must respond to late in the \needlegame{} game, and Fig.~\ref{fig:nm-levels} shows a wider range of the levels used to collect training data.

\subsubsection{Task Representation}

Our goal is to reproduce user skills in a way that closely approximates the human performances of these skills, but applies them to complex novel environments.
The Needle Master task is to move from left to right, hitting gates in a specific order while attempting to avoid the dark red areas on the top and bottom of the gate if possible.
The needle in this game has a state defined $s = [\tt{x}, \tt{y}, \theta]^T$, where $\tt{x}$ and $\tt{y}$ denote the position of the needle and $\theta$ represents its orientation. It is associated with an action $u = [\tt{u}, \tt{v}]$. The dynamics of the needle are given by:

\myeq{needle-dynamics}{
	\begin{bmatrix}\tt{x}_{i+1} \\ \tt{y}_{i+1} \\ \theta_{i+1} \end{bmatrix} = 
	\begin{bmatrix} \tt{x} + \tt{v} \cos(\theta) \\ \tt{y} + \tt{v} \sin(\theta) \\ \theta + \tt{u} \end{bmatrix}
}

We can represent a trajectory produced by these dynamics as a series of curves with constant rotation $\tt{u}$, constant velocity $\tt{v}$, and duration $\tt{t}$, similar to that described in~\cite{kobilarov2011cem}.

We segment task demonstrations based on a set of predicates.
These predicates describe the relationship of the needle to its environment, as a proxy for what the expert demonstrating the trajectory is responding to.
Predicates used in these examples are:
\begin{itemize}
  \item \texttt{NEEDLE-IN-GATE}: true if the needle is within a gate
  \item \texttt{HAS-PREV-GATE}: true if we are leaving one gate and still have gates to visit
  \item \texttt{GATE-CLOSED}: true if a gate has been closed
  \item \texttt{GATE-OPEN}: true if a gate is open still
  \item \texttt{AT-EXIT}: true if the needle has moved off the screen to the right, thus ending the level
\end{itemize}

These predicates divide user demonstrations into five different states, parameterized by different entities in the environment: (1) approaching the first gate, (2) passing through a gate, (3) connecting two gates, (4) moving to the exit, and (5) at the exit and done with a level.
There are features associated with each entity in the environment. Different actions use whichever features are appropriate: the action for connecting two gates uses the features associated with both gates, for example, while passing through a gate is based on the features associated with that gate alone.

Each gate is defined by its position, height, width, and its orientation. Top and bottom bars are a constant fraction of the gate's height.
The features for each gate project the needle's current position into the gate's frame of reference, and compare the distance between the angle of the needle and the angle of the gate:
\myeq{gate}{
\phi^{gate}(s) = 
\begin{bmatrix}
\cos\left(\text{atan2}\left(\tt{y},\tt{x}\right) + \theta^{gate}\right) - \tt{x}^{gate} \\
\sin\left(\text{atan2}\left(\tt{y},\tt{x}\right) + \theta^{gate}\right) - \tt{y}^{gate} \\
\sqrt{(\tt{x} - \tt{x}^{gate})^2 + (\tt{y} - \tt{y}^{gate})^2} \\
|\theta^{gate} - \theta| \\
\end{bmatrix}
}

We use the $y$ distance to the edge of the level as a feature for skills learned after the $\texttt{GATE-CLOSED}$ predicate is true for all gates, and we use the magnitude of the rotation control $|\tt{u}|$ as a feature for all actions.
Finally, we use the change in the above variables as an additional set of features. In addition, we describe whether or not the needle is in tissue with a boolean variable.

We used GMMs with $k=3$ components to model skills and fit them through the Expectation-Maximization algorithm.
These were learned from three expert demonstrations from each of the first twelve levels of the game. 
Our approach allows us to complete very complex Needle Master levels including those with obstacles and changing constraints due to presence of tissue, as shown in Fig.~\ref{fig:nm-levels}(top).

\subsubsection{Comparison To Other Methods}

To validate our approach, we showed how our algorithm could generate trajectories for $10$ different randomly generated levels with multiple gates and obstacles.
We compare these results to two other approaches: (1) a naive version of the algorithm without performing planning,
and (2) a version of the algorithm that does include the goal term discussed in Eq.~\myeqref{weights-task}.

All three versions of the algorithm rely on the same GMM learned from user data.
In the ``Naive'' version of the algorithm, at each time step we take the action corresponding to the highest probability by searching for the optimal $(\mathbf{u},\mathbf{v})$ given the current features $x$.

\begin{table}[bt!]
\centering
\begin{tabular}{| p{1.4cm} | p{1.2cm} | p{1.2cm} | p{1.2cm} | p{1.4cm} |}
\hline
  & Gates Entered & Gates Cleared & Gates Broken & Levels Finished \\
\hline \hline
Naive & 4 & 0 & 4  & 10 \\
No Goals & 9 & 2 & 7 & 5 \\
Full & 20 & 16 & 4 & 10\\
\hline
\end{tabular}
\caption{Comparison of performance on the Needle Master task with different methods implemented. 
}
\label{table:nm-comparison}
\end{table}

We compare four different metrics: (1) how many gates were ever entered by the needle, (2) of those, how many gates were ``cleared'' successfully, (3) how many gates were ``broken'' by the needle either touching the top and bottom bars, and (4) how many levels were successfully finished (final position off the screen to the right).
In one set of experiments (top), we generated 10 levels with two gates and two obstacles. In the second (bottom) set, we generated 10 levels with two gates and no obstacles.

Table~\ref{table:nm-comparison} shows the results of these experiments.
The naive approach performed very poorly, usually because it has trouble aligning with the gates. 
When gates are in unexpected positions, it choses high-probability movements that end up moving it away from the correct path. It finished every level, though, since it chooses the most likely motion and in all examples we move from left to right.
The approach without a goal term similarly performs poorly: the behavior of each component action is not well defined at the beginning and end.


\subsection{Case Study: Robotic Tasks}\label{sec:robots}

Our algorithm is demonstrated in an object manipulation task where we need to build a structure of increasing complexity out of magnetic blocks, as per the task described in~\cite{bohren2013pilot}.
This simulated assembly task consists of a number of skills which need to be executed in the same order to connect two pieces together.
In the structure assembly task, we are learning the actions:

\begin{itemize}
\item \texttt{APPROACH(link)}: move to a ``link'' object
\item \texttt{GRASP(link)}: close the gripper and secure the link object.
\item \texttt{ALIGN(link,node)}: holding the link, move so that it is above the node.
\item \texttt{PLACE(link, node)}: place the link on top of the node.
\item \texttt{RELEASE(link)}: open the gripper
\item \texttt{DISENGAGE(link)}: move away from the link without knocking it over
\end{itemize}

\begin{figure*}[bt!]
\includegraphics[width=2\columnwidth]{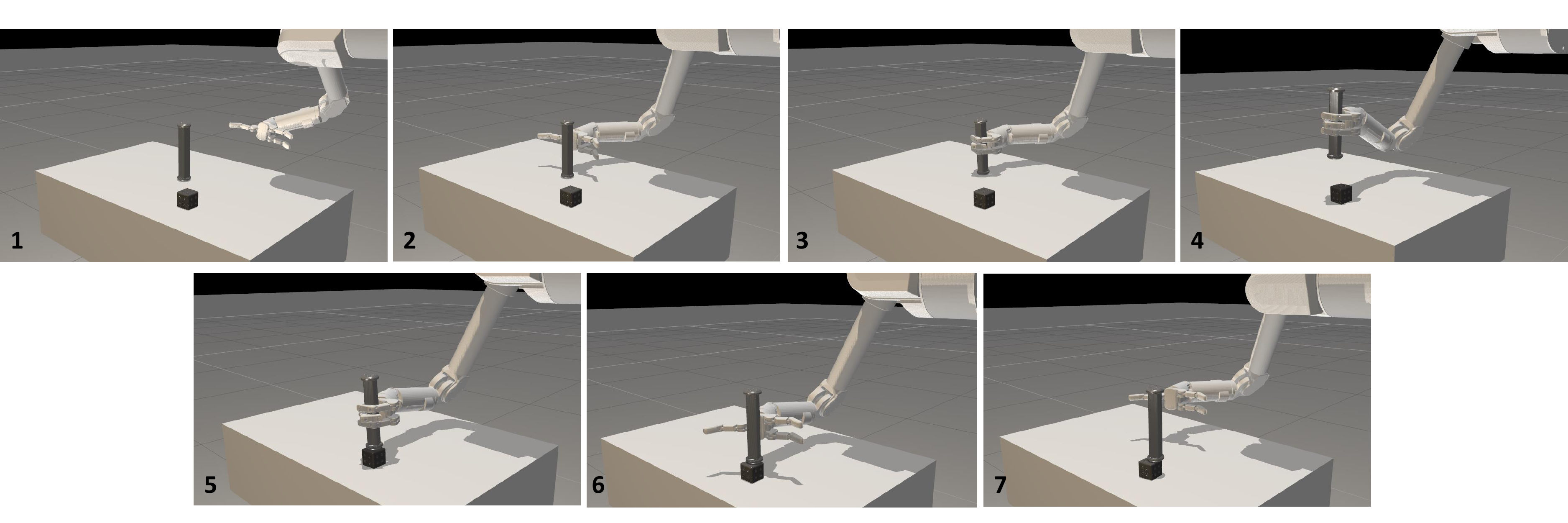}
\caption{Sequence of actions necessary to perform the structure assembly task. Starting in the upper left: initial state, \texttt{APPROACH(link)}, \texttt{GRASP(link)}, \texttt{ALIGN(link,node)}, \texttt{PLACE(link,node)}, \texttt{RELEASE(link)}, and \texttt{DISENGAGE(link)}.}
\label{fig:assembly}
\end{figure*}

These skills must be chained together in a specific order, and each skill is critical to the completion of the next skill. Figure~\ref{fig:assembly} shows the procedure as executed by our algorithm.
We also consider the task of grasping and lifting a sander above a vise in an industrial robot ``workshop assistant'' task.

When teaching the \texttt{APPROACH} and \texttt{GRASP} skills, we are teaching one particular grasp, associated with a range of valid positions. Note that the \texttt{link} object has eight-way rotational symmetry: we can flip it over or rotate in 90 degree increments and end up with a valid grasp. We can also mirror our grasp, putting two fingers on either side of the object.
The same is true of the \texttt{ALIGN} and \texttt{PLACE} actions for connecting the link and the node: there are four possible mates for each of the cube's six surfaces. We are teaching one specific mate and grasp in this example, but we could use these symmetries to apply our grasp and mate to any of the applicable surfaces on our two objects.

\begin{figure}[bt!]
\includegraphics[width=\columnwidth]{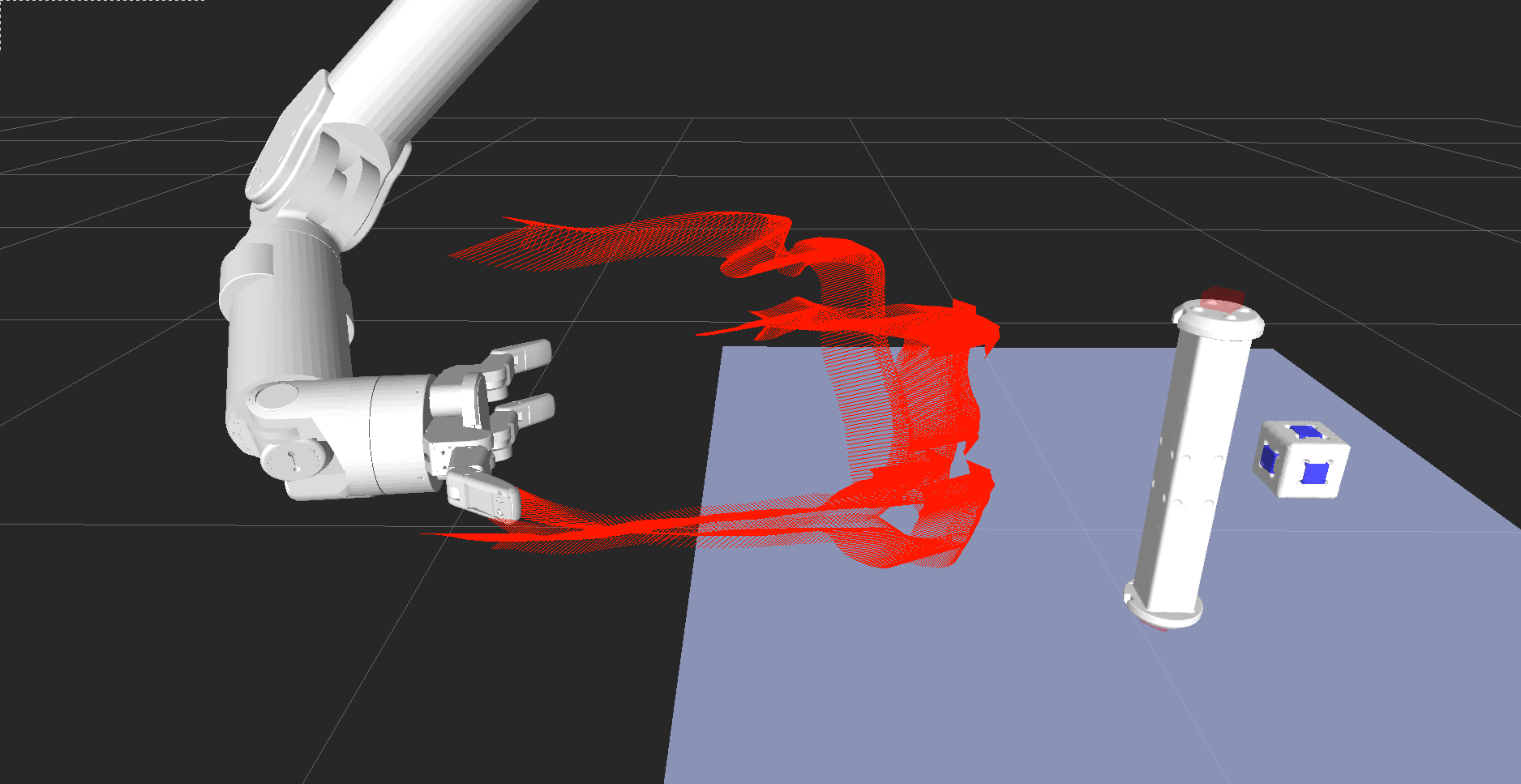}
\caption{Four trajectories showing user demonstrations used to train the \texttt{APPROACH} skill for the simulated Barrett WAM arm task. Note that these trajectories capture a very wide range of possible approaches, being of different lengths and in different regions.}
\label{fig:training}
\end{figure}

To create a model of each of these skills, we collect four demonstrations of the object manipulation action starting from the same grasp, with two of the Barrett Hand's fingers on the left side of the link and one on the right. The simulated WAM arm was teleoperated: users controlled the end effector frame of the robot, and joint torques were computed by a Jacobian Transpose Null Space controller.
As a result training data was very noisy, as we might expect from similar non-expert instruction of a real robot. Fig.~\ref{fig:training} shows the trajectories used to train one of these skill models, that of the \texttt{APPROACH} action. Trajectories are shown in red, relative to the \texttt{link} object that parameterizes the skill.

\subsubsection{Task Representation}


We use four sets of features: (1) the time in a particular state, (2) the gripper command variables, (3) the transforms between the end frame and the objects.
We used the $\tt{x}$,$\tt{y}$, and $\tt{z}$ offsets from this transform, the distance (norm of the position), and the quaternion representation of the rotation as features in the transform.

The feature function $\phi(s)$ computes the forward kinematics of the robot arm: our variables are the relative end effector position in comparison to objects in the environment and the norm of the position.
For our experiments, we selected the relevant objects in each demonstration for which we compute the features.
For the \texttt{APPROACH}, \texttt{GRASP}, \texttt{RELEASE}, and \texttt{DISENGAGE} actions, these are computed based on the \texttt{link} object alone. For the \texttt{ALIGN} and \texttt{PLACE} actions, we compute parameters based on the \texttt{node} object.

We represent a robot trajectory as a set of end-effector poses to attain.
In the WAM arm simulation, we use three such poses, and interpolate between them to create trajectories.
Each individual primitive consists of an $(\tt{x},\tt{y},\tt{z},\tt{roll},\tt{pitch},\tt{yaw})$ offset from the current end effector location. We interpolate between these points to create a trajectory.

\subsection{Results}\label{ssec:comparison}

\begin{figure}[bt!]
  \centering
  \includegraphics[width=\columnwidth]{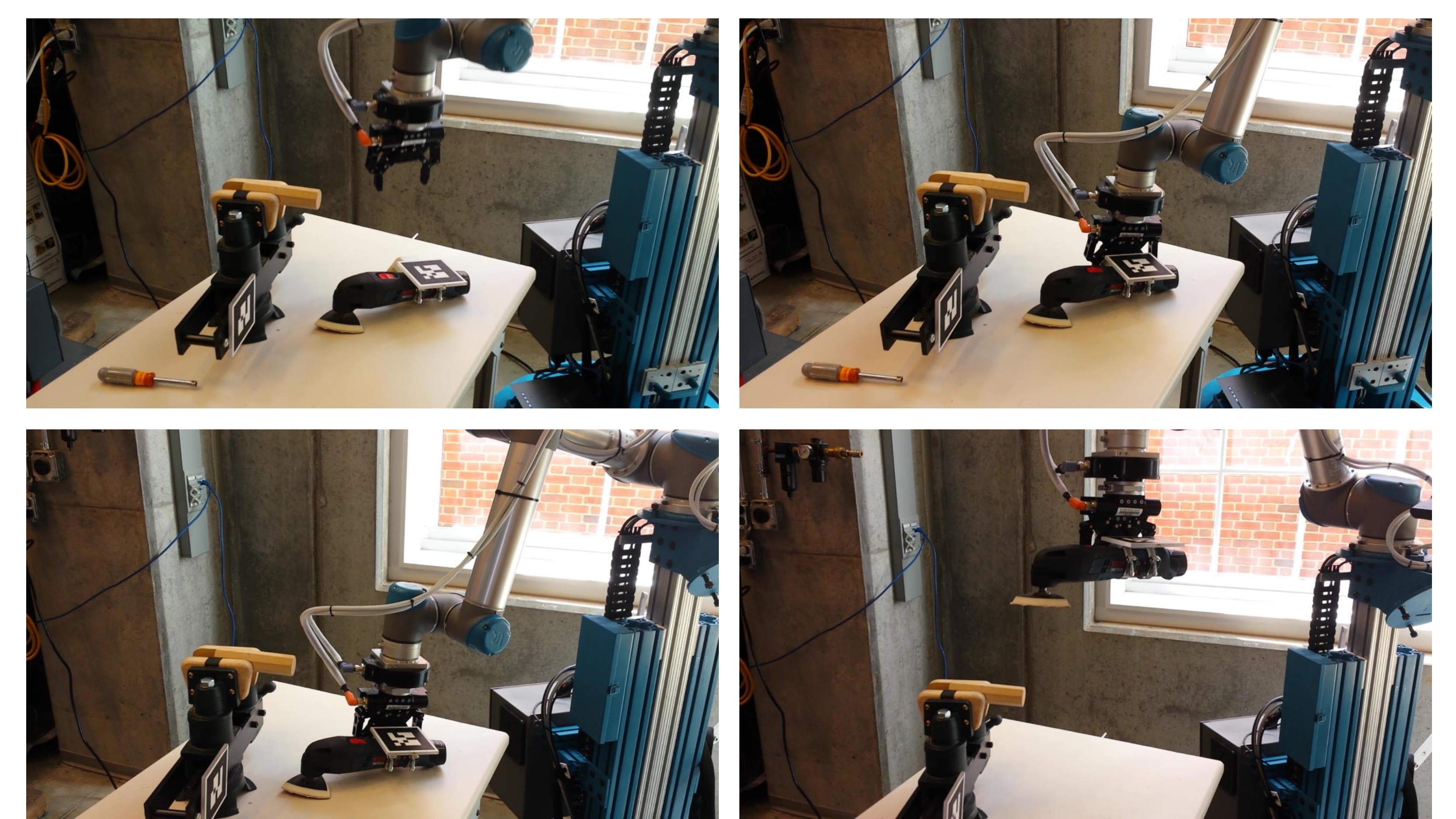}
  \includegraphics[width=\columnwidth]{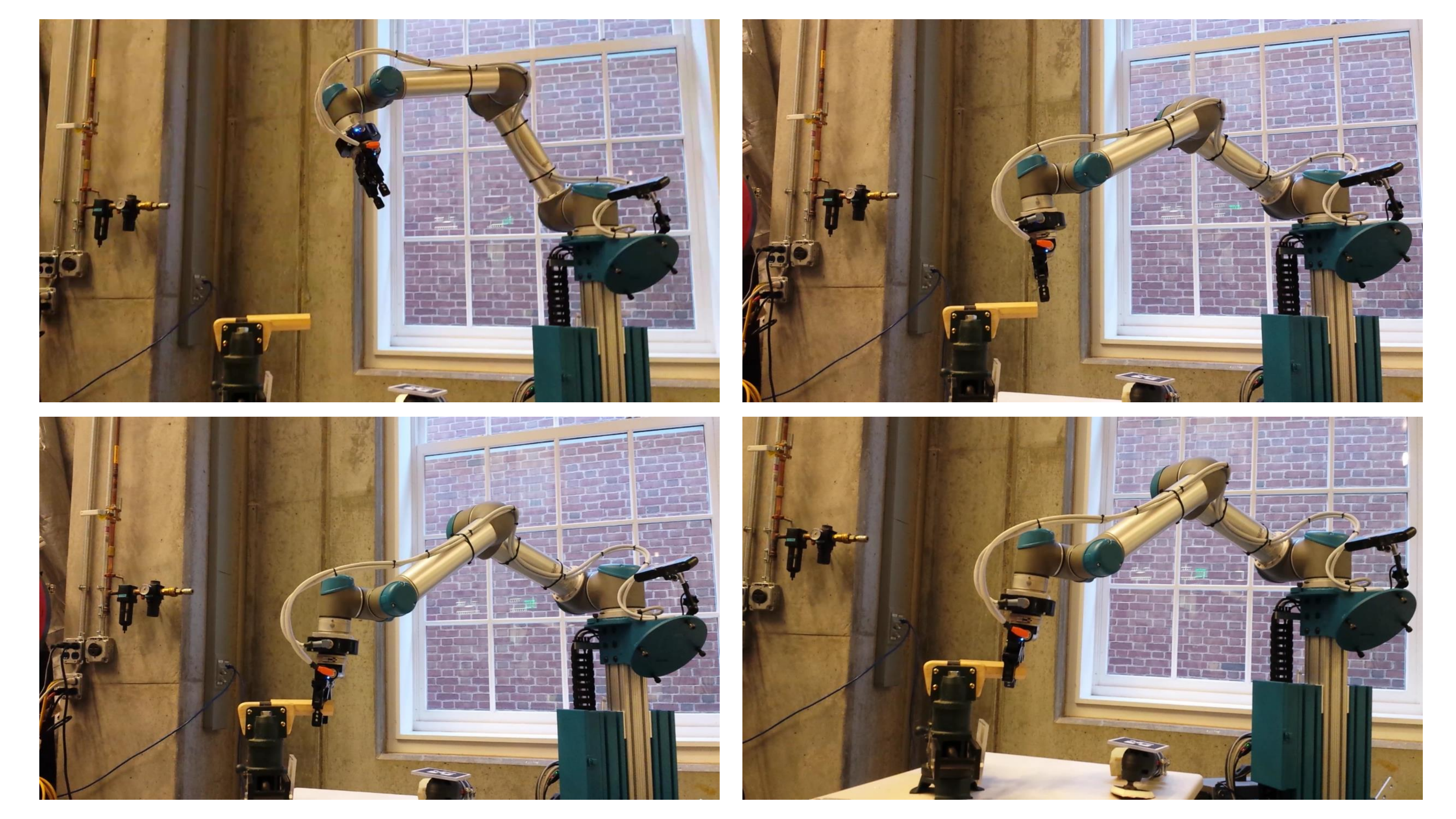}
  \caption{Workshop assistant experiments. Top: the first portion of a sanding task, wherein the robot grabs a sander and lifts it up over the vise. Bottom: a grasp action, moving to pick a wooden block from a vise.}
  \label{fig:workshop}
\end{figure}

We collected three demonstrations of each of the different skills with a dynamic simulation of the Barrett WAM arm.
We then place these pieces in different positions in the environment, and validated our method by performing the task in different locations.
The results of one performance in a novel environment are shown in Fig.~\ref{fig:assembly}.
We also successfully demonstrated the task on a simple tool-use example, grabbing and lifting a sander after being given a human demonstration of the same. A performance of this task with a new world configuration is shown in Fig.~\ref{fig:workshop}.

\begin{figure}[bt!]
  \centering
  \includegraphics[width=0.8\columnwidth]{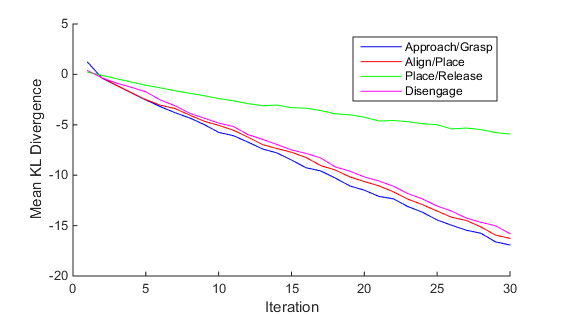}
  \caption{Plot showing improvement of mean KL divergence over 30 iterations of the algorithm with $\alpha=0.75$.}
  \label{fig:kl}
\end{figure}

\begin{figure}[bt!]
  \centering
  \includegraphics[width=0.8\columnwidth]{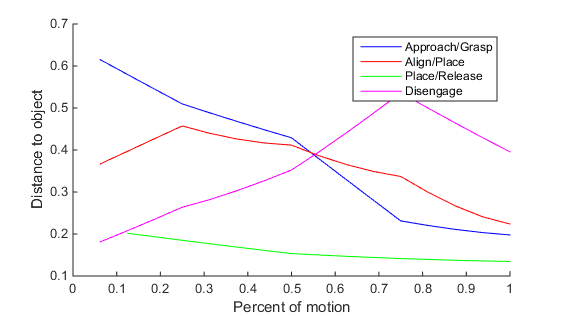}
  \caption{Plot showing distance between the end frame and relevant objects for learned skills. For \texttt{APPROACH} and \texttt{DISENGAGE}, this object is the link. For \texttt{ALIGN} and \texttt{PLACE} this is the node.}
  \label{fig:dists}
\end{figure}

In particular, we find that as in the Needle Master example discussed above, it is very important to have good models of the next skill in order to complete each previous skill: motion in the \texttt{PLACE} and \texttt{APPROACH} actions are often very noisy, but successful grasps mates (as exemplified by the \texttt{GRASP} and \texttt{RELEASE} skills) are much more restrictive.

Skills such as \texttt{APPROACH} and \texttt{ALIGN} represent vaguely defined actions, as well.
In addition, the training data used to create these skills are extremely noisy due to the wide variety of different approach directions and distances from which a user can move.
One advantage of our approach is that these actions become better defined when combined with a following action, like the \texttt{GRASP} or \texttt{PLACE}.

Fig.~\ref{fig:kl} shows how our approach iteratively decreases the KL divergence between the features produced by the current trajectory distribution parameterized by $v_i$ and the expert probability distribution. Each line indicates the average KL divergence during iteration for one action (either \texttt{APPROACH}, \texttt{ALIGN}, \texttt{PLACE}, or \texttt{DISENGAGE}) and its goal.
The distribution for the action \texttt{ALIGN} with the goal of \texttt{PLACE} is far less constrained, since there are many ways of approaching the region above the node before a mate.
Fig.~\ref{fig:dists} explores how the distance between various objects changes over the course of these trajectories. We can see that during the \texttt{ALIGN} action, distance does not change very much, partly because much of the action is lifting the link up so that it can be mated to the top. Contrast this with the \texttt{APPROACH} action, where the end effector quickly approaches the link and shifts into a high-quality position to transition to \texttt{GRASP}.

\section{Conclusions}

We described a practical approach for motion planning based on a probabilistic model of a skill.
One of the goals of this approach is to enable task-level planning: 
by representing skills as probability distributions, we create a framework that allows us to describe and combine a broad range of cost functions. We validated this approach with a series of experiments in different application domains.

Future work will integrate this approach with higher-level task planning.
In addition, we will examine automatically segmenting demonstrations of tasks into sub-skills and identifying useful invariant features.
In particular, we could use Bayesian non-parametric methods such as those discussed in~\cite{grollman2010incremental,butterfield2010learning,niekum2012learning} in order to automatically determine skills and transitions between skills from user data. This would eliminate the requirement that the algorithm must be provided with labeled training data.

\bibliographystyle{IEEEtran}
\bibliography{paper}

\end{document}